\pdfoutput=1

\documentclass[11pt]{article}

\usepackage[]{EMNLP2022}

\usepackage{times}
\usepackage{latexsym}
\usepackage{xcolor}
\usepackage[T1]{fontenc}

\usepackage[utf8]{inputenc}

\usepackage{microtype}

\usepackage{inconsolata}

\usepackage{times}
\usepackage{latexsym}
\usepackage{multirow}
\usepackage{subfig}
\usepackage[title]{appendix}
\usepackage{color}
\usepackage[utf8]{inputenc}
\usepackage{amsmath}
\usepackage{amssymb}
\usepackage{algpseudocode}
\usepackage{algorithm}
\usepackage{paralist}
\usepackage{makecell}
\usepackage{multirow}
\usepackage{tabularx}
\usepackage{booktabs}
\usepackage{bbding}
\usepackage{graphicx}

\usepackage[T1]{fontenc}

\usepackage[utf8]{inputenc}

\usepackage{microtype}

\title{PcMSP: A Dataset for Scientific Action Graphs Extraction from Polycrystalline Materials Synthesis Procedure Text}

\author{Xianjun Yang$^{1}$, Ya Zhuo$^{2}$, Julia Zuo$^{2}$, Xinlu Zhang$^{1}$, Stephen Wilson$^{2}$, Linda Petzold$^{1}$\\
	\small$^1$Department of Computer Science
	\small$^2$Department of Materials Science and Engineering\\
        \small University of California, Santa Barbara \\
	{\small \tt \{xianjunyang, yzhuo, jlzuo, xinluzhang, stephendwilson, petzold\}@ucsb.edu}\\
}

\begin{document}
\maketitle
\begin{abstract}
Scientific action graphs extraction from materials synthesis procedures is important for reproducible research, machine automation, and material prediction. But the lack of annotated data has hindered progress in this field. We demonstrate an effort to annotate \textbf{P}oly\textbf{c}rystalline \textbf{M}aterials \textbf{S}ynthesis \textbf{P}rocedures (\textbf{PcMSP}) from 305 open access scientific articles for the construction of synthesis action graphs. This is a new dataset for material science information extraction that simultaneously contains the synthesis sentences extracted from the experimental paragraphs, as well as the entity mentions and intra-sentence relations. A two-step human annotation and inter-annotator agreement study guarantee the high quality of the PcMSP corpus. We introduce four natural language processing tasks: sentence classification, named entity recognition, relation classification, and joint extraction of entities and relations. Comprehensive experiments validate the effectiveness of several state-of-the-art models for these challenges while leaving large space for improvement. We also perform the error analysis and point out some unique challenges that require further investigation. We will release our annotation scheme, the corpus, and codes to the research community to alleviate the scarcity of labeled data in this domain\footnote{https://github.com/Xianjun-Yang/PcMSP}.

\end{abstract}

\section{Introduction}
Synthesis procedural texts are written in instructional languages \citep{grishman2001adaptive, grishman2014analyzing} to represent the step-by-step reactions, but also contain the distinct features in specific domains, such as the domain notations, writing styles, and journal requirements. The synthesis procedures of materials science articles include valuable information for new materials prediction \citep{raccuglia2016machine}, laboratory automation \citep{coley2019robotic} and knowledge graph construction \citep{mrdjenovich2020propnet}. However, available datasets are extremely limited, despite the notable work by \citep{mysore2017automatically, mysore2019materials, friedrich2020sofc, o2021ms}.

\begin{table}[]
    \centering
    \scalebox{0.9}{
    \begin{tabular}{p{7.7cm}} \toprule
        \textbf{Synthesis Paragraph} \\\midrule
        \\ ${\color{brown}\textit{Polycrystalline}_{\texttt{[Descriptor]}}}$ sample of composition ${\color{red}\textit{Sr2CoO4}_{\texttt{[Material\_target]}}}$ was ${\color{orange}\textit{synthesized}_{\texttt{[operation]}}}$ under ${\color{violet}\textit{high pressure}_{\texttt{[Property\_pressure]}}}$ at ${\color{teal}\textit{ high temperature}_{\texttt{[Property\_temperature]}}}$. Starting materials of ${\color{blue}\textit{SrO2}_{\texttt{[Material\_recipe]}}}$ and ${\color{blue}\textit{Co}_{\texttt{[Material\_recipe]}}}$ were  ${\color{brown}\textit{well}_{\texttt{[Descriptor]}}}$ ${\color{orange}\textit{mixed}_{\texttt{[operation]}}}$ in a ${\color{brown}\textit{molar ratio}_{\texttt{[Descriptor]}}}$ of ${\color{blue}\textit{SrO2}_{\texttt{[Material\_recipe]}}}$ : ${\color{blue}\textit{Co}_{\texttt{[Material\_recipe]}}}$=${\color{pink}\textit{2 : 1}_{\texttt{[Value]}}}$. The ${\color{yellow}\textit{mixture}_{\texttt{[Material-intermedium]}}}$ was ${\color{orange}\textit{sealed}_{\texttt{[operation]}}}$ into ${\color{pink}\textit{a}_{\texttt{[Value]}}}$ ${\color{brown}\textit{gold}_{\texttt{[Descriptor]}}}$ ${\color{purple}\textit{capsule}_{\texttt{[Device]}}}$. 
        ...
        The crystal structure of the polycrystalline sample was identified by the powder X-ray diffraction (XRD, Rigaku Smart- lab3), using Cu-K$\alpha$ radiation ($\lambda$=1.54184Å). ...
        \\ \bottomrule
    \end{tabular}}
    \caption{An example of a synthesis paragraph from our dataset with index srep27712 \citep{li2016magnetization}.}
    \label{tab:paragraph}
    \vspace{-0.24cm}
\end{table}

\begin{figure}[htbp]
\centering
\includegraphics[width=\linewidth]{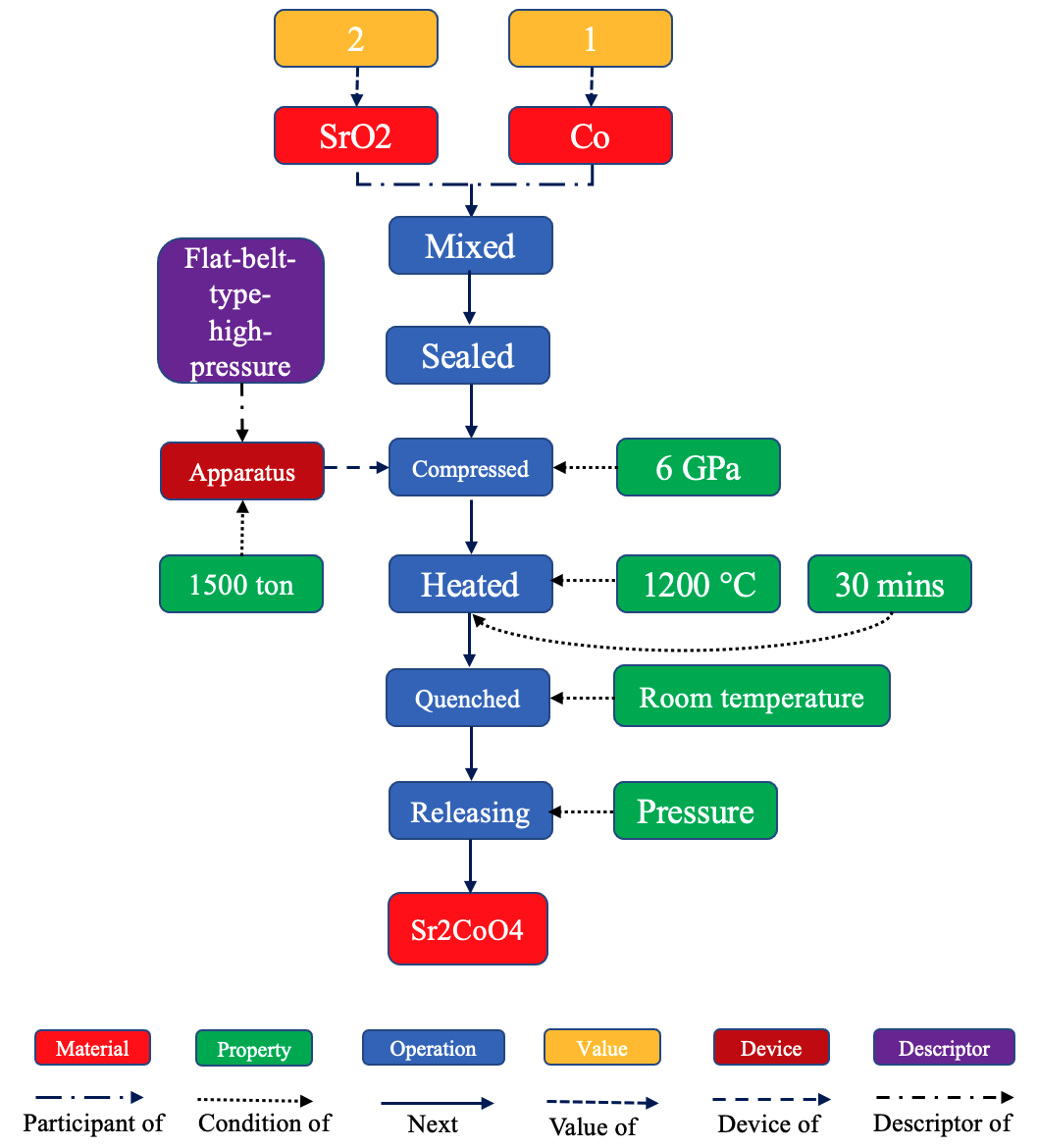} 
\caption{A synthesis action graph constructed from Table~\ref{tab:paragraph}.}
\label{figure0}
\end{figure}

The goal of information extraction from procedures is to construct the action graphs, which refer to all the steps in a synthesis making up a Directed Acyclic Graph (DAG) \citep{mysore2019materials, kulkarni2018annotated} (as can be seen from one example in Figure~\ref{figure0}). This can be further breakdown into three tasks: sentence classification, named entity recognition (NER), and relation extraction (RE). Previous research \citep{mysore2017automatically, mysore2019materials} either annotates the whole synthesis paragraph in the general inorganic domain, ignoring the non-synthesis sentences and subdomain discrepancy or only focuses on entity mentions \citep{friedrich2020sofc, o2021ms}. 

\begin{figure*}[htbp]
\centering
\includegraphics[width=\linewidth]{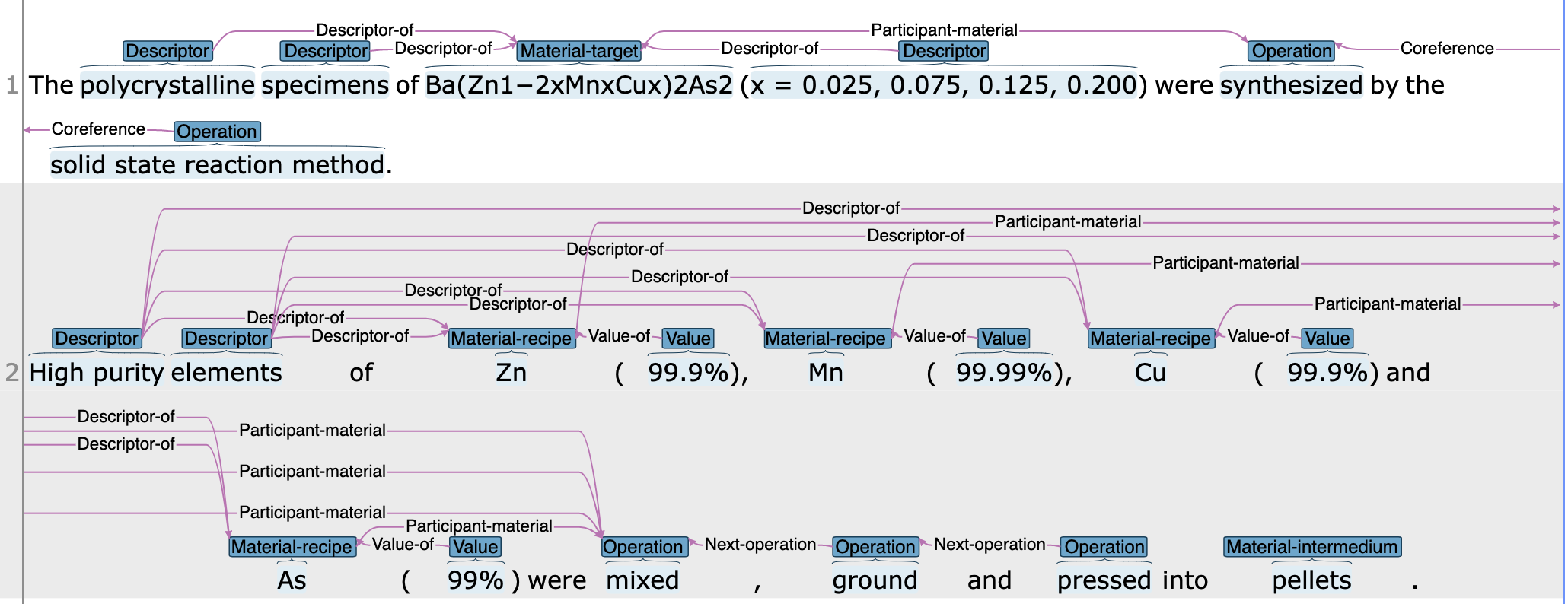}
\caption{An annotated PcMSP example on the INCEpTION platform, taken from srep15507 \citep{man2015ba}.}
\label{figure1}
\end{figure*}

To fill this gap, we focus on one important category of polycrystalline materials and simultaneously include all three tasks. The annotation guidelines are designed by materials experts after comprehensive discussion, and the new dataset is subsequently labeled with a two-round annotation.

The key contributions of this paper include:
\begin{itemize}
\item We contribute a new large-scale dataset, as well as an annotation scheme with high quality for information extraction in materials science. 
\item We conduct comprehensive experiments on four tasks, sentence classification, named entity recognition, relation extraction, and joint extraction to provide baselines.
\item We perform error analysis and point out unique challenges and potential use of this dataset for future research.

\end{itemize}
\section{Related Work}\label{related}

\subsubsection*{\textbf{Scientific information extraction}}
With the fast-growing volume of scholarly publications, it is highly demanding to extract structured information from large-scale scientific literature in many domains \citep{augenstein2017semeval, luan2018multi, jiang2019role, beltagy2019scibert, buscaldi2019mining}, like biomedical domain \citep{shah2003information, lai2021joint, zhang2021fine, lewis2020pretrained, kulkarni2018annotated} and chemistry domain \citep{rocktaschel2012chemspot, he2020overview}. In the field of materials science, there have been few attempts in this direction, leaving many unexplored challenges for research \citep{hong2021challenges}. Recent research mainly focuses on knowledge base construction \citep{jiang2019role, luan2018multi}, new materials discovery \citep{isayev2019text}, and automation of lab procedures \citep{vaucher2020automated, tamari2021process, steiner2019organic}. \citep{beltagy2019scibert} trained a Bidirectional Encoder Representations from Transformers model (SciBERT) on 1.14M scientific papers from Semantic Scholar for scientific information extraction. 

\subsubsection*{\textbf{Materials procedures information extraction}}
In the area of annotation of materials synthesis procedures, \citep{mysore2019materials} annotate 230 general materials synthesis paragraphs for NER and RE tasks. Similar work is also undertaken by \citep{friedrich2020sofc}, in which 45 open access scholarly articles are labeled for experiment-describing sentence classification, NER, and slot filling tasks. However, in contrast to our works, their annotation scheme focuses on the full text rather than the experimental section. \citep{kuniyoshi2020annotating} annotate the synthesis process of all-solid-state batteries from the scientific literature, but their corpus is not publicly available. \citep{walker2021impact} release MatBERT trained on 50 million materials science paragraphs to explore the impact of domain-specific pre-training on NER task. Also of interest, \citep{o2021ms} recently create the largest corpus for entity mentions extraction in both general domain and subdomain from material synthesis text, but the relations between entities are still missing. 

\subsubsection*{\textbf{Named entity recognition and relation extraction}}
Many neural network-based models have been proposed for named entity recognition, for example, \citep{huang2015bidirectional, lample2016neural, panchendrarajan2018bidirectional}. The core idea uses one encoding layer (e.g. Long  Short-Term Memory (LSTM) \citep{hochreiter1997long}, BERT) for representation and one additional conditional random fields (CRF \citep{Lafferty2001ConditionalRF}) layer for sequence labeling. Then relations are predicted based on either gold entities or predicted entities, and PURE \citep{zhong2021frustratingly} designs two separate encoders for joint extraction of entities and relations. We adopt their model for our tasks due to its super performance.

\section{The Selection of Our Dataset}
Here we talk about the importance of our selection and how is it different from other materials procedural text corpora. 

\textbf{Why do we choose inorganic polycrystalline materials?} 
There are a number of sub-categories within solid-state inorganic materials. For example, materials can be divided based on function and properties, such as the battery or thermoelectric materials. Synthesis within both categories largely falls within the broader category of solid-state synthesis and even then, there is a high degree of overlap with other function categories, such as quantum and magnetic materials. More importantly, \textbf{those materials are usually in the form of polycrystalline}. Other subcategories relate to form factors, for instance, single-crystalline synthesis often starts with a \textbf{polycrystalline} synthesis and therefore has a high degree of overlap with solid-state synthesis. 

Inorganic polycrystal compounds span combinations of the entire periodic table and different chemical bonding schemes, such that their synthesis typically takes place under extreme conditions, such as high temperature and pressure. Reaction pathways are therefore difficult to characterize without specialized equipment and are not well established for any given material. In particular, solid-state reactions, which are the main techniques to synthesize inorganic polycrystalline materials, are particularly similar to a “black box”, where materials scientists can only make educated guesses to the procedure or stability of a new reaction. This presents a prime opportunity \citep{mysore2017automatically, mysore2019materials} for compiling published inorganic synthesis data in order to demystify the black box of solid-state inorganic materials synthesis and create datasets for future text mining endeavors. While there have been efforts within general solid-state materials \citep{mysore2017automatically, mysore2019materials, o2021ms} and battery materials subcategory \citep{friedrich2020sofc}, this work aims to extend the subcategory of inorganic solid-state synthesis methods in order to address the frequent overlap and “borrowing” of materials between subdisciplines of materials science.

\textbf{Why do we discard characterization sentences?}
Inorganic reactions typically involve relatively few reactions from a set of precursors and there are very few purification pathways for solid materials compared to organic materials or liquids. Therefore, characterizations of solid-state inorganic reactions are seldom reported in literature unless they proceed to complete purity within standard measurement fidelity. This is in contrast to organic materials where there are a number of important characterization metrics in a compound, such as molecular weight in polymers or reaction yield. 
Therefore, these standard characterization measurements do not add valuable information for a researcher attempting to recreate the reported synthesis method and we decide to discard these characterization sentences.

\textbf{Why do we annotate sentence, entity, and relation simultaneously?}
A full action graph consists of both entities and relations extracted from experimental-describing sentences. However, most previous research either ignores the annotation of sentence or relation information, making them incomplete for action graph construction. To fill this gap, we aim to annotate all pertinent information jointly.

\section{Description of the Annotation}\label{method}
\subsection{Selection of synthesis procedures for annotation } \label{probdefin}
We begin by harvesting the polycrystalline materials synthesis-related open access publications from the main journal publishers by searching keywords (e.g. 'polycrystalline+synthesis'). The journals that we used include Physical Review Journals\footnote{https://journals.aps.org/}, Nature journals\footnote{https://www.nature.com/}, Science journals\footnote{https://www.science.org/journals}, Journal of the American Chemical Society\footnote{https://pubs.acs.org/journal/jacsat}, Advanced Materials\footnote{https://onlinelibrary.wiley.com/journal/15214095}, Journal of Physics Condensed Matter\footnote{https://iopscience.iop.org/journal/0953-8984}, Chemistry of Materials\footnote{https://pubs.acs.org/journal/cmatex} and ArXiv\footnote{https://arxiv.org/}. After the collection of 305 publications, each portable document format (PDF) document is converted into a plain text file by pdfminer\footnote{https://pdfminersix.readthedocs.io/en/latest/}. The experimental paragraphs usually appear in the experimental section within an article and are selected by one materials expert. To improve the data quality, the selected paragraphs are double-checked by another annotator to ensure their correctness. And some missing sentences caused by the conversion process are also added. Finally, the collected paragraphs are prepared for the next step of annotations.

\begin{table*}
\centering
\scalebox{0.65}
{
\begin{tabular}{c|c|c|c|c|c|c|c|c|c}
\textbf{Dataset} & \textbf{Domain} & \textbf{Procedure only} & \textbf{Documents} & \textbf{Sentence type} &
\textbf{Sentences} & \textbf{Entity type} & \textbf{Entities} & \textbf{Relation type} & \textbf{Relations} \\
\toprule
MSPT & General & \XSolidBrush & 230 & \XSolidBrush & 2112 & 21 & 20849 & 16 & 18402\\
SOFC-Exp & Subdomain & \XSolidBrush & 45 & 2 & 853 & 16 & 5095 & \XSolidBrush & \XSolidBrush\\
SC-CoMIcs & Subdomain& \XSolidBrush & 1000 & \XSolidBrush & 6639 & 7 & 42337 & \XSolidBrush & \XSolidBrush\\
MS-MENTIONS & General & - & 595 & \XSolidBrush & 7980 & 14 & 44295 & \XSolidBrush & \XSolidBrush\\
\hline
\textbf{Our PcMSP} & Subdomain & \Checkmark & 305 & 2 & 2468 & 13 & 14592 & 8 & 13968\\
\bottomrule
\end{tabular}
}
\caption{Corpus statistics of our PcMSP and previous datasets for materials science. \XSolidBrush denotes that no such information is contained in the corresponding corpus. - denotes that the corpus has not been released yet.}
\label{tab:datacomparison}
\end{table*}

\subsection{Sentence annotation}\label{sec:Sentence}
Based on the selected paragraphs from the aforementioned step, each document is annotated on the semantic annotation platform INCEpTION \citep{tubiblio106270}, and the sentence segmentation is carried out automatically\footnote{InCeption uses Java’s built-in sentence segmentation algorithm with US locale.}. Each line represents all tokens of one sentence, and the annotation is done on the token level. In practice, only the synthesis-related sentences are annotated for NER and RC. The resulting unlabeled sentences automatically obtain non-synthesis labels. This process resulted in 1497 synthesis-related sentences and 971 non-related sentences. It is worthwhile to point out that several selected paragraphs also contain single crystal synthesis (this occurs < 1\%), but we do not take those as synthesis-related sentences so as to focus purely on polycrystalline synthesis. In general, most non-synthesis sentences are relevant to the characterization of materials, description of devices, etc. While synthesis sentences typically describe the synthesis actions conducted in the experiments. For example, in Table~\ref{tab:paragraph}, the first two sentences are synthesis-related while the remaining sentences are not.

\subsection{Entity type annotation}\label{sec:NER annotation}
We defined 13 entity types to include the most useful entity mentions, which are decided by the materials experts. Each span of continuous words is labeled as a certain kind of entity type. There are five general categories of labels, namely \textbf{Material}: Material-target, Material-recipe, Material-intermedium and Material-others, \textbf{Property}: Property-time, Property-temperature, Property-rate and  Property-pressure, \textbf{Operation}, \textbf{Item}: Value, Brand, Device and \textbf{Descriptor}. Every general coarse-grained category can further be divided into one or several fine-grained types. The full definitions of these labels can be found in the following. \\
\textbf{Material-target}: final material (or products) of the material synthesis process, usually refers to only one target in a typical procedural paragraph, but can appear as multiple target materials (this occurs less than 1\%).  \\
\textbf{Material-recipe}: raw material used to synthesize the final product, can be fundamental elements(like $Si$), compounds(like $SrO2$), or precursors of other polycrystalline materials. \\
\textbf{Material-intermedium}: an intermediate material produced during the synthesis process that is subsequently used as participants in the following reactions. \\
\textbf{Material-others}: materials that are not compositionally related to the final material or used as solvents (like $water$) to provide reaction conditions.\\
\textbf{Operation}:  an individual action performed by the experimenters, which is often represented by verbs or a particular overall synthesis method, like $Solid-state-reaction$. \\
\textbf{Property-time}: a time condition associated with an operation, which is usually composed of numerical values and time units. \\
\textbf{Property-temperature}: a temperature condition associated with an operation, which is usually composed of numerical values and temperature units. \\
\textbf{Property-rate}: a rate condition associated with an operation, which is usually composed of numerical values and rate units. The rates can be rotation speed, cooling,  or heating rates, etc. \\
\textbf{Property-pressure}: a pressure condition associated with an operation, which is not only in the form of value and units but also can be a certain condition like vacuum, helium, or air. \\
\textbf{Value}: numerical values and their corresponding units. In addition, we include specifications like "around", "over", “more than” or “between” in the annotation span (e.g., “around 250 g,” and “over 20 mol”). We do not include time, temperature, pressure, or rate in this category, as they are already included in properties. \\
\textbf{Device}: mentions of the type of device used in the corresponding operation, which can contain the device name and serial number.\\
\textbf{Brand}: the brand name or source laboratory associated with the equipment or material. \\
\textbf{Descriptor}: description of an operation or a material or a value that does not apply to properties but is necessarily included for clear descriptions.\\

\subsection{Relation type annotation}\label{sec:REA}
The previous two steps provide us with the labeled entity mentions within each sentence. We then connect each entity pair by a relation type when there is a believed necessary connection, according to the definition of agreement study. The full descriptions of relation labels are listed in the following.\\
\textbf{Participant-material}: materials that are involved in one operation process, and we also mark the target material and its synthesis action as this label.\\
\textbf{Device-of-operation}: a device used in an operation.\\
\textbf{Condition-of}: indicates the conditions of an operation (such as the temperature, time, and pressure) for performing an operation. \\
\textbf{Value-of}: expresses the relationships between participated material and their weight, mass, volume, or purity, and also represents the relationship between the device and its serial number. \\
\textbf{Next-operation}: represents the order of an operation sequence that one operation that happens following the previous operation. Note that we assume the linear sequence of synthesis operations happens sentence by sentence, which is true for most cases. \\
\textbf{Brand-of}: expresses the relationships between a raw material or device and its manufacturer name or source laboratory.\\
\textbf{Descriptor-of}: the descriptor for the material, device, or operation that can not be covered by other labels. \\
\textbf{Coreference}: represents the same material or operation in the same sentence. 

Besides, according to the largest Document-level relation extraction dataset \citep{yao2019docred}, around 40\% of relations exist across multiple sentences. But cross-sentence relation is out of our scope for current work and we leave it for future investigation.

\section{Inter-annotator Agreement Study}\label{app:inter1}
We perform a two-round agreement study to ensure that our corpus has a high quality of annotation. Before undertaking the formal annotation, all four annotators participate in a discussion of the formulation rules and discuss the necessary entity and relation types. In the warm-up exercise, all annotators annotate the same documents individually and then compare and discuss the results together to achieve better agreement on annotation. After the agreements are formulated, in the first-round annotation four annotators are randomly assigned different documents to work on. It takes around twenty to thirty minutes to annotate one document on average for all annotators. When all of the annotations are finished, two of the four annotators select several typical examples for analysis and eventually set more rules for annotating the most debatable parts. In the second round of annotation, two lead annotators individually re-annotate half of the documents, guaranteeing that there are no significant differences or mistakes. It takes around 500 hours for our material expert team in total to create this corpus to guarantee high quality.

We use Fleiss’ Kappa to measure the agreement scores between our four annotators. The result is shown in Table~\ref{tab:Agreement}, with substantially high agreement scores. We can see obvious improvements in all aspects from the first to second round annotation, demonstrating the effectiveness of our annotation pipeline. We use five metrics to measure the agreement score: Sen. refers to sentence agreement, En1. means span boundaries and type are both correct, En2. means matched type on same spans, Re1. represents complete relation triple with correct entities and Re2. stands for correct relation type on same entities. 
More details are discussed in Appendix~\ref{app:inter}. 

\begin{table}
\centering
\scalebox{0.8}
{
\begin{tabular}{c|c|c|c|c|c}
\textbf{Round} & \textbf{Sen.} & \textbf{En1.} & \textbf{En2.} & \textbf{Re1.} & \textbf{Re2.}\\
\toprule
{First-round} & 80.13 & 56.41 & 92.8 & 48.51 & 90.2\\
{Second-round} & 85.06 &69.81 & 93.44 & 53.63 & 91.03\\
\bottomrule
\end{tabular}
}
\caption{Two-round inter-annotator agreement study measured by Fleiss’ Kappa.}
\label{tab:Agreement}
\end{table}

\section{Statistics of Corpus and Problem Formulation}\label{sec:Dataset}
In this section, we describe the statistics of this new dataset, the comparison with precious corpora, and formulated tasks. 
\subsection{\textbf{PcMSP corpus}}
We outline the main material science corpus in Table~\ref{tab:datacomparison}, including Materials Science Procedural Text (MSPT) \citep{mysore2019materials}, SC-CoMIcs \citep{yamaguchi2020sc}, SOFC-exp \citep{friedrich2020sofc} and MS-MENTIONS \citep{o2021ms}, as well as our PcMSP corpus. Among those corpora, MSPT focuses on general solid-state compounds and is most similar to ours. But MSPT contains annotation for all sentences in synthesis procedural paragraphs, even though many of those sentences are actually describing material characteristic methods rather than synthesis procedures. 
On the other hand, the SC-CoMIcs and MS-MENTIONS only contain entity mentions, without any sentence or relation labels. In addition, the SOFC-exp corpus focuses on the whole articles rather than the procedural text and does not contain full annotation of entity-to-entity relations. The provided relations in the original SOFC-exp dataset are constructed by only linking slot fillers to the syntactically closest EXPERIMENT mention.

Our new PcMSP dataset simultaneously contains the sentence, entity, and relation annotation from 305 polycrystalline synthesis-related open access publications. Among the 2468 sentences extracted from the synthesis paragraphs, 1497 sentences are identified as the synthesis description involved in an experiment. A total of 14608 entity mentions with 13 entity types and 13987 relations with 8 relation types are labeled by materials experts. We further show more corpus statistics for the training, validation, and test set in Table~\ref{tab:Statistics}. We provide the train/validation/test split for potential use in the future.
\subsection{\textbf{Task definition}}
The PcMSP corpus labels every sentence with entity mentions and relations among entity pairs. Formally, given a sentence of n words $s = \{w_1, ..., w_n\}$ with the labeled sentence type, entity set $\mathcal{E}$ and relation set $\mathcal{R}$, four information extraction tasks are introduced:

1) SC: classification of the sentence as an experimental procedure sentence or irrelevant sentences, 2) NER: recognition of all named entities mentions in $\mathcal{E}$, 3) RE: identification of the entity pair relations in $\mathcal{R}$ and 4) Joint: joint extraction of all entities and relations.
\begin{table}
\centering
\scalebox{0.9}
{
\begin{tabular}{c|c|c|c}
\textbf{Item} & \textbf{Train} & \textbf{Validation} & \textbf{Test}\\
\toprule
{Synthesis procedures} & {243} &{31} & 31\\
{Sentences} & {1972} &{275} & 221\\
{Avg. sentence length}  & {27.24} &{26.22} & 27.21\\
{Avg. sentences/Doc} & {8.12} &{8.87} & 7.13 \\
\hline
{Entities} & {11585} & {1507} & 1516 \\
{Entity types}  & {13} & {13} &  13\\
{Relations} & {11176} & {1376} & 1435 \\
{Relation types} & {8} & {8} & 8\\
{Tokens}  & {53720} &{7210} & 6014\\
\bottomrule
\end{tabular}
}
\caption{Statistics of our annotated dataset.}
\label{tab:Statistics}
\end{table}

\section{Results and Analysis}\label{experiment_set}
\begin{table}
\centering
\scalebox{0.85}
{
\begin{tabular}{ c|c | c c c} 
\multirow{1}{*}{\makecell{ }}
&  \multicolumn{1}{c| }{ \textbf{ Dev} }  
&  \multicolumn{3}{c  }{ \textbf{Test} }\\
\toprule
Model & F1 & P & R & F1(\%) \\
\hline 
BERT-base & 87.84 & 89.43 & 85.92 & 87.20 \\
SciBERT & 88.38 & 89.84 & 88.12 & 88.85 \\
MatBERT & 89.44 & \textbf{91.71} & 89.13 & 90.16 \\
\hline
Human evaluation & - & 90.74 & \textbf{90.62} & \textbf{90.62} \\
\bottomrule
\end{tabular}
}
\caption{Experiment-describing sentence classification results in terms of F1 score on the test set. Scores are reported on macro average.}
\label{tab:Experiment detection}
\end{table} 

We present the main experimental results in this section, and more modeling details are included in Appendix~\ref{app:model}. PURE refers to the advanced joint extraction model by \citep{zhong2021frustratingly}. For all the experiments, we use the \emph{bert-base-uncased} \citep{devlin2019bert}, \emph{scibert-scivocab-uncased} \citep{beltagy2019scibert}, and \emph{matbert-base-uncased} \citep{walker2021impact} as encoders. Generally, BERT with domain-specific pretraining considerably improves the performance.

\subsection{Sentence classification}
We summarize the results for the experiment-describing sentence detection in Table~\ref{tab:Experiment detection}. For this binary classification task, we fine-tune the BERT, SciBERT, and MatBERT \citep{walker2021impact} models, resulting in an F1 score of 87.20, 88.85, and 90.16\%, respectively. The best result is achieved by MatBERT, demonstrating the usefulness of domain-specific pretraining.
The close-human performance of sentence classification stems from the obvious difference in expression between synthesis-describing sentences and others. Generally, synthesis-describing sentences contain 1) the material's chemical formulas, 2) the operations (usually certain verbs), and 3) experimental conditions. In contrast, other sentences often describe the characterization approaches which are totally different. In conclusion, synthesis sentence detection is the foundation for other downstream tasks and the high detection accuracy guarantees the success of our workflow for other downstream tasks.

\begin{table}
\centering
\scalebox{0.8}
{
\begin{tabular}{ c|c | c c c} 
\multirow{1}{*}{\makecell{ }}
&  \multicolumn{1}{c| }{ \textbf{ Dev} }  
&  \multicolumn{3}{c  }{ \textbf{Test} }\\
\toprule
Model & F1 & P & R & F1(\%) \\
\hline
BERT + PURE &77.06 & 79.23 & 77.24 & 78.23\\
MatBERT + PURE &76.98 & \textbf{79.56} & \textbf{79.36} & \textbf{79.46}\\
\hline
SciBERT + PcMSP & 79.46 & 77.32 & {78.91} & 78.84\\
\qquad + MS-Mentions & 91.55 & - & - & 91.47\\
\qquad + MSPT & 82.8 & - & - & 78.15\\
\qquad + SOFC-Exp & 73 & -&- & 78.57\\
\hline
{Human evaluation} & - & \textbf{90.05} & \textbf{89.26} & \textbf{89.46} \\ 
\bottomrule
\end{tabular}
}
\caption{Named entity recognition results in terms of F1 score on the PcMSP test set.}
\label{tab:NER1}
\end{table} 
\subsection{Named entity recognition}
In Table~\ref{tab:NER1}, we present the NER results obtained from different models. Based on the synthesis procedure sentences detected earlier, we train the models only on the experiment-describing sentences, ignoring irrelevant sentences. The SciBERT model is trained with one CRF layer for sequence labeling and the MatBERT is stacked with one additional forward layer for span-based tagging.
The MatBERT model with PURE achieves the best F1 result of 79.46\%, although a large gap of 10 points still exists compared with the human agreement score. When looking at all the label performance from Table~\ref{tab:NER2}, recognizing the labels such as $Property-rate$, $Property-time$ and $Operation$ achieves good scores of 92.31\%, 84.38\%, and 83.39\%, respectively. On the contrary, the recognition is still difficult for labels like $Material-others, Material-interdium$, etc. One possible reason might be those mentions require cross-sentence reasoning, while the current model is only trained on single sentences. We also report SciBERT results on other previously mentioned materials procedural datasets and the overall sentence-level results are very consistent. Thus, a promising direction for improving the results is to include paragraph-level context or use cross-domain transfer learning and we leave this for future work.

\begin{table}
\centering
\scalebox{0.75}
{
\begin{tabular}{c|c|c|c|c}
\textbf{ Entity Label} & \textbf{Number} & \textbf{P} & \textbf{R} & \textbf{F1} \\
\toprule
$Brand$ & 21 & 66.67 & 80.00 & 72.73 \\ 
$Descriptor$ & 324 & 61.34 & 74.30 & 67.20 \\
$Device$ & 79 & 66.67 & 79.37 & 72.46 \\
$Material-intermedium$ & 96 & 55.68 & 50.52 & 52.97 \\
$Material-others$ & 27 & 1.00 & 16.67 & 28.57 \\ 
$Material-recipe$ & 150 & 70.66 & 75.16 & 72.84 \\
$Material-target$ & 65 & 67.74 & 68.85 & 68.29  \\  
$Operation$ & 329 & 82.30 & 84.51 & 83.39 \\
$Property-pressure$ & 41 & 62.22 & 70.00 & 65.88 \\   
$Property-rate$ & 15 & 92.31 & 92.31 & 92.31  \\
$Property-temperature$ & 77 & 76.74 & 79.52 & 78.11  \\
$Property-time$ & 72 & 83.08 & 85.71 & 84.38 \\
$Value$ & 187 & 76.63 & 87.58 & 81.74  \\
\hline
Overall & 1483 & \textbf{77.32} & \textbf{78.91} & \textbf{78.84} \\ 
\hline
{Human evaluation} & - & \textbf{90.05} & \textbf{89.26} & \textbf{89.46} \\ 
\bottomrule
\end{tabular}
}
\caption{NER per label performance on the PcMSP test set by SciBERT.}
\label{tab:NER2}
\end{table}

\subsection{Relation classification}
In this section, the modeling is performed on gold entities to investigate individual modeling capability.
The relation classification results are provided in Table~\ref{tab:RE2}. For entity pairs without any relation, a ‘NA’ label is given for modeling. Here, the human agreement score is calculated by treating one annotation as gold and another one as predictions. Among all of the relation modeling results in Table~\ref{tab:RE2}, we can see that the F1 score is almost always above 80\%, demonstrating promising prediction results on all label levels. In particular, the $Condition-of$ and $Brand-of$ relation predictions achieve a high F1 score of 89.21\% and 88.46\%, respectively. But $Coreference$ prediction is more difficult, achieving only 71.74 points. Overall, the RE modeling achieves comparable results to those of human annotators, although leaving more than 10\% points for improvement. Similarly, we believe cross-sentence information can further improve the results and leave it for further investigation.

\begin{table}
\scalebox{0.7}
{
\begin{tabular}{c |c |c |c |c }
\textbf{ Relation Label} & \textbf{Number} & \textbf{P} & \textbf{R} & \textbf{F1} \\
\toprule
$Brand-of$ & 25 & 85.19 & 92.00 & 88.46 \\ 
$Condition-of$ & 212 & 90.73 & 87.74 & 89.21 \\
$Coreference$ & 140 & 72.79 & 70.71 & 71.74 \\
$Descriptor-of$ & 349 & 83.92 & 88.25 & 86.03 \\
$Device-of-operation$ & 87 & 86.59 & 81.61 & 84.02 \\ 
$Next-operation$ & 109 & 84.62 & 90.83 & 87.61 \\
$Participant-material$ & 296 & 80.74 & 80.74 & 80.74  \\  
$Value-of$ & 217 & 87.67 & 88.48 & 88.07 \\
$NA$ & 7102 & 97.62 & 97.42 & 97.52 \\
\hline
Overall & 8534 & \textbf{85.54} & \textbf{86.42} & \textbf{85.93} \\ 
\hline
{Human evaluation} & - & \textbf{96.82} & \textbf{97.69} & \textbf{97.37} \\ 
\bottomrule
\end{tabular}
}
\caption{RE per label performance on the PcMSP test set.}
\label{tab:RE2}
\end{table}

\subsection{Joint entity and relation extraction}
Previous sections consider entity and relation extraction separately, but the practical scenario involves joint extraction of entities and relations. Here we use the super performing joint extraction PURE \citep{zhong2021frustratingly} model to evaluate the joint extraction performance. The PURE model first produces all the possible entities and then uses these predicted entities for relation extraction. Following their work, the evaluation is conducted on three metrics: (1) \textbf{Ent}: a predicted entity is correct only if the predicted span boundaries and entity type are both correct. (2) \textbf{Rel}: a predicted relation type is correct given the correct boundaries of two spans. (3) \textbf{Rel+}: in addition to the boundaries requirements, the predicted entity must conserve the correct type.

As can be seen from Table~\ref{tab:joint}, the joint model demonstrates a 79.46\% F1 score in terms of the entity prediction. As for the relation prediction, a much lower F1 score is observed for both Rel and Rel+, with 66.69\% and 62.53\% respectively. This is not unexpected since the RE relies on the previous entity prediction result and the error inevitably propagates. Compared with previous individual extraction, the joint extraction achieves lower results and leaves a large margin for improvement. Considering the goal of action graphs extraction from procedures is the joint extraction of all entities and relations, we encourage more research towards better modeling. Also of notice, the current joint evaluation is on a single sentence, while more realistic end-to-end extraction is conducted on the whole paragraph. And cross-sentence relations will also preserve in such a scenario, but this is out of the scope of this work.
\begin{table}
\centering
\scalebox{1.0}
{
\begin{tabular}{ c | c c c} 
\textbf{Joint} & \textbf{P} & \textbf{R} & \textbf{F1} (\%) \\
\toprule
Ent & 79.56 & 79.35 & 79.46 \\
Rel & 67.55 & 65.85 & 66.69 \\
Rel+ & 63.33 & 61.74 & 62.53 \\
\bottomrule
\end{tabular}
}
\caption{Joint entity and relation extraction results on test set. }
\label{tab:joint}
\end{table} 

\section{Conclusion}
In summary, we contribute a new dataset PcMSP collected from 305 open access scholarly publications for action graphs construction from material synthesis procedures. The two-round human expert's annotations guarantee the high quality of the dataset, which is evident by the agreement study. Based on this new dataset, we perform sentence classification, named entity recognition, and relation extraction tasks. We also experiment with the joint extraction of entities and relations.
Several good-performing neural models are utilized to provide competitive baselines, although leaving a big gap compared with the human upper bound. To alleviate the data scarcity of this domain, we will make our dataset publicly available. 

Some future directions would be to investigate incorporating cross-sentence context, improving the joint extraction results, performing paragraph-level end-to-end extraction, as well as using our PcMSP to investigate domain adaptation. For example, pre-training with distant supervision in the materials domain might also help improve the results. Considering the high labeling cost, how to efficiently transfer knowledge into other domains to reduce human annotations is also of great importance.

\section*{Limitations}
Even though we try our best to guarantee high annotation quality, inaccurate labels may still exist. We are not responsible for any products derived from our dataset.
Also, the real-world end-to-end actions graphs construction involves the whole pipeline and will inevitably face the error propagation problem. 

\section*{Ethics Statement}
We notice that our data source comes from open access publications and we make our dataset publicly available, but further use might also fall into potential limitations required by certain journals. Besides, in our annotation process, all the annotators are paid as research assistants following the campus policy. 

\section*{Acknowledgements}
We thank the anonymous ARR and EMNLP reviewers for their insightful comments related to this paper.
We thank the insightful suggestions from Lei Li for an early version of the manuscript.
We gratefully acknowledge support from the UC Santa Barbara NSF Quantum Foundry funded via the Q-AMASEi program under NSF award DMR-1906325. Any opinion or conclusions expressed in this material are those of the author(s) and do not necessarily reflect the views of the National Science Foundation.


\bibliography{anthology,custom}
\bibliographystyle{acl_natbib}

\appendix

\section{Background on Polycrystalline Materials}
\label{sec:Appendix}
Polycrystalline materials are solids composed of small randomly oriented crystallites, also called grains, with the size varying from a few nanometers to several millimeters. Most of the inorganic solid materials available in macroscopic quantities are in fact polycrystals, including common metals, ceramics, and rocks. They provide versatility in numerous applications such as superconductors, batteries, photovoltaic cells, and shape memory alloys \citep{husain2018review, peng2018key, peng2017review, biswal2021recent}.

\begin{figure}[htbp]
\includegraphics[width=\linewidth]{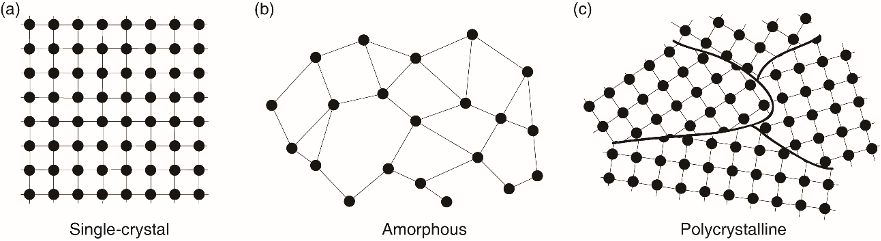}
\caption{Material classification based on the degree of atomic order: (a) single-crystal, (b) amorphous, (c) polycrystalline.}
\label{outline}
\end{figure}

The structure of a single crystal or monocrystal (Figure 3a) is continuous and highly ordered, while an amorphous phase (non-crystal) (Figure 3b) such as glass does not display any structures, as the constituent atoms are not arranged in an ordered manner. In-between these two extremes, a polycrystal (Figure 3c) exists, which is made up of many crystallites, also referred to as grains. During the solidification of polycrystalline materials, small nuclei first form at different spots of the liquid sample and subsequently absorb atoms from the surrounding liquid to grow into larger grains. These grains vary in size from nanometers to millimeters and are randomly oriented with no preferred direction in the structure. Therefore, a large enough volume of polycrystalline material can be approximately considered isotropic. Compared to single crystals, polycrystalline materials also require less sophisticated techniques to make, significantly lowering the cost of production. As most real-world solids are polycrystalline materials, it is critical to synthesize and understand polycrystalline materials. A substantial number of studies have been done by researchers across the world to discover new materials. This work exacts knowledge from those synthesis processes and aims to guide the synthesis efforts toward the unexplored space.

\section{Modeling}\label{app:model}
We mainly use PURE \citep{zhong2021frustratingly} as backbones for our tasks. 
\subsection{Sentence classification}
Sentence classification is a binary text classification problem. We build one additional layer on top of the BERT and fine-tune it for another 10 epochs. 

\subsection{Named entity recognition}
For the SciBERT model, we stack another conditional random field (CRF) \citep{tseng2005conditional} layer on top of SciBERT for sequence labeling following the traditional BIO notation.
For the MatBERT result, we follow the span-based approach in \citep{zhong2021frustratingly} to obtain the contextualized representation for any span and feed it into another forward layer to predict the entity type.

\subsection{Relation classification}
We utilize the span representations of entity mentions for relation prediction with typed entity markers as proposed by the relation model in \citep{zhong2021frustratingly}. 
 
\subsection{Joint extraction}
Following \citep{zhong2021frustratingly}, the predicted entities are fed into another encoder for relation prediction. And we adopt two different encoders for the joint extraction of entities and relations. 

\section{Experimental settings}
We select the best combination of hyperparameters from the development set by random search. Three random seeds are chosen for all models, and we report the results based on the median performance. The standard macro-average precision(\textbf{P}), recall(\textbf{R}), and \textbf{F1} scores are calculated.

The Adam optimizer \cite{Kingma2015AdamAM} is used for all models. Other parameters are selected within a range of values, for example learning rate ranges from [1e-4, 5e-5, 1e-6] and batch size of 8 or 16. 
The models are implemented in PyTorch\footnote{https://pytorch.org}, and a Tesla P40 with 24GB RAM is used for all experiments. The model takes around half-hour, one hour, and three hours for the training of sentence, entity, and relation tasks for 10 to 50 epochs.

\subsection{Data preprocessing}\label{sec:process}
Each plain text document containing the synthesis paragraphs is imported into the INCEpTION platform, which also performs the sentence segmentation and word tokenization by its built-in algorithm. After tokenization, each sentence is mapped with the corresponding entity mentions and relations, which includes the named entity type, position, token information, and the relations type, as well as left and right position information. 

\section{Inter-annotator Agreement Study}\label{app:inter}
Despite from Fleiss' kappa for measuring agreements in Table~\ref{tab:Agreement}, we describe more details in this section.

\subsection{Sentences annotation}
Given a paragraph selected from a scientific publication, we first examine the synthesis-related sentences. In practice, the annotators only label synthesis-related sentences for the entity and relation information. All other sentences without labeling are considered non-synthesis sentences. To compare the model’s performance with human annotation, 32 documents are labeled by two main annotators in the second round individually. Then one annotation is regarded as the ground truth and the other is treated as a prediction. A micro-average F1 score of 90.62\% is calculated between the two annotators. Additional details about the precision, recall, and F1 score is shown in Table~\ref{tab:sen}. In general, the main annotator selects 153 of the 256 sentences to label as synthesis-related sentences, while the second annotator chose 163 to be labeled as target sentences. The overall result demonstrates high-quality annotations and can serve as a human agreement score for further baseline.
\begin{table}
\centering
\scalebox{0.9}
{
\begin{tabular}{c|c|c|c|c}
\textbf{Sentence Label} & \textbf{Number} & \textbf{P} & \textbf{R} & \textbf{F1}\\
\toprule
Synthesis & 153 & 89.57 & 95.42 & 92.41 \\ 
Non-synthesis & 103 & 92.47 & 83.50 & 87.76 \\
\hline
Overall & 256 & 90.74 & 90.62 & 90.62\\
\bottomrule
\end{tabular}
}
\caption{Human agreement score on experiment-describing sentences.}
\label{tab:sen}
\end{table}
\subsection{Named entity annotation}
Following the previous step, all of the entity mention boundaries are first recognized by the annotators and then one entity label is chosen from the predefined entity labels to represent the entity type. Among the recognized overlap of 143 experiment-describing sentences from the previous step by both annotators, one annotator recognizes 1483 named entities while the second annotator considers 1345 entity mentions as necessary to be labeled. The agreement metric is calculated by treating one result as the true value, while the second result is used as a predictive value. The overall P, R, and F1 scores are given in Table~\ref{tab:NER} in terms of per label performance. As can be seen from the results, two of the annotators agreed on the majority of the labels, while in some circumstances (like $Material-others$), the score is relatively lower, due possibly to a different understanding of those entity mentions.
\begin{table}
\centering
\scalebox{0.7}
{
\begin{tabular}{c|c|c|c|c}
\textbf{Entity\_Label} & \textbf{Number} & \textbf{P} & \textbf{R} & \textbf{F1}\\
\toprule
$Brand$ & 21 & 94.74 & 85.71 & 90.00 \\ 
$Descriptor$ & 324 & 83.49 & 82.72 & 93.10 \\
$Device$ & 79 & 93.67 & 93.67 & 93.67 \\
$Material-intermedium$ & 96 & 87.37 & 86.46 & 86.91 \\
$Material-others$ & 27 & 74.19 & 85.19 & 79.31 \\ 
$Material-recipe$ & 150 & 86.84 & 88.00 & 87.42 \\
$Material-target$ & 65 & 96.83 & 93.85 & 95.31  \\  
$Operation$ & 329 & 94.08 & 91.79 & 92.92 \\
$Property-pressure$ & 41 & 90.00 & 87.80 & 88.89 \\   
$Property-rate$ & 15 & 92.86 & 86.67 & 89.66  \\
$Property-temperature$ & 77 & 86.59 & 92.21 & 89.31  \\
$Property-time$ & 72 & 95.71 & 93.06 & 94.37 \\
$Value$ & 187 & 91.57 & 87.17 & 89.32  \\
\hline
Overall & 1483 & \textbf{90.05} & \textbf{89.26} & \textbf{89.46} \\ 
\bottomrule
\end{tabular}
}
\caption{Human agreement score on NER.}
\label{tab:NER}
\end{table}

\subsection{Relation annotation}
Here we focus on relation annotation based on a given entity pair. When both annotators first agree on the same entity pair, the agreement F1 score is 97.37\%, demonstrating the high quality of the annotation.
\begin{table}
\centering
\scalebox{0.7}
{
\label{tab:NER }
\begin{tabular}{c|c|c|c|c}
\textbf{Entity\_Label} & \textbf{Number} & \textbf{P} & \textbf{R} & \textbf{F1}\\
\toprule
$Brand-of$ & 18 & 100.0 & 100.0 & 100.0 \\ 
$Condition-of$ & 174 & 100.0 & 97.13 & 98.54 \\
$Coreference$ & 69 & 81.43 & 82.61 & 82.01 \\
$Descriptor-of$ & 256 & 93.94 & 96.88 & 95.38 \\
$Device-of-operation$ & 69 & 98.48 & 94.20 & 96.30 \\ 
$Next-operation$ & 99 & 98.97 & 96.97 & 97.96 \\
$Participant-material$ & 229 & 94.35 & 94.76 & 94.55 \\  
$Value-of $ & 162 & 97.53 & 97.53 & 97.53 \\
\hline
Overall & 1076 & \textbf{96.82} & \textbf{97.69} & \textbf{97.37} \\ 
\bottomrule
\end{tabular}
}
\caption{Human agreement score on RC.}
\end{table}

\begin{figure}[ht]
\centering
\includegraphics[width=\linewidth]{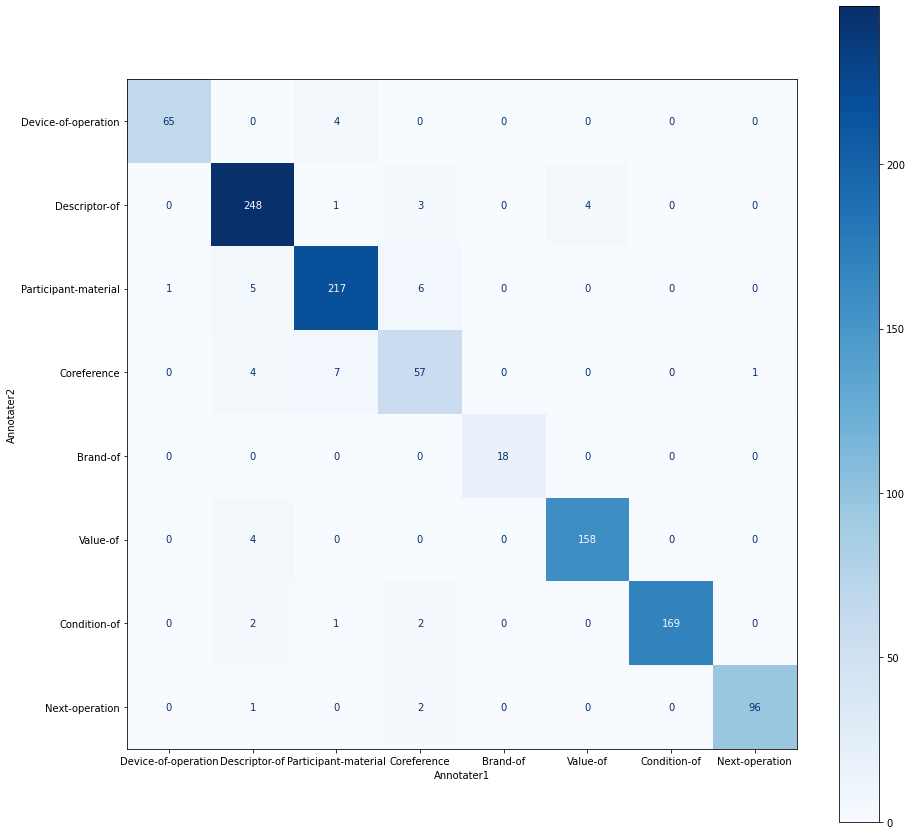}
\caption{Confusion matrix over relations between the two lead annotators.}
\label{cm1}
\end{figure}

Figure~\ref{cm1} shows the confusion matrix of relations between the two lead annotators.

\begin{table}
\centering
\begin{tabular}{c|c|c|c}
\textbf{Journal} & \textbf{Train} & \textbf{Validation} & \textbf{Test}\\
\toprule
Elsevier & 46 & 6 & 4 \\
ArXiv & 81 & 5 & 8 \\ 
Nature family & 71 & 13 & 13 \\ 
ACS family & 13 & 4 & 2 \\ 
APS family & 28 & 3 & 4 \\ 
Others & 4 & 0 & 0 \\ 
\bottomrule
\end{tabular}
\caption{Document distribution among main journals: ACS: American Chemistry Society, APS: American Physical Society, and others refers to other journals not included here.}
\label{tab:journals}
\end{table}

\section{Document Distribution Among Journals}
Table~\ref{tab:journals} demonstrates that the source of our collected documents is distributed among different journals. Considering that the writing style and publication requirements of different journals vary a lot, we aim to include documents from a range of sources to make the dataset more diverse.

\section{Annotation Examples and Statistics}\label{app:er}
Common examples of entity mentions and relation triples are shown in Table~\ref{tab:NERs} and Table~\ref{tab:REs}, respectively. The relation triple has the form of $r_i$: ($e_i$, $e_j$), where $r_i$ is one relation label, while $e_i$ and $e_j$ denote the entity mention within one sentence.
\begin{table*}
\centering
\begin{tabular}{c|c|c|c|c}
\textbf{ Entity Label} & \textbf{Count} & \textbf{Frequent mentions} & \textbf{Percentage}\\
\toprule
$Descriptor$ & 2450 & Polycrystalline, quartz, polycrystalline & 21.15 \\
$Material-target$ & 442 & Ca2CeCr2TiO9, powder, sample & 3.82  \\ 
$Brand$ & 317 & Alfa Aesar, Aldrich, Sigma-Aldrich & 2.74 \\ 
$Device$ & 662 & tube, crucible, glove box & 5.71 \\
$Material-intermedium$ & 772 & pellets, mixture, samples & 6.66 \\
$Material-others$ & 158 & water, distilled water, oxygen & 1.36 \\ 
$Material-recipe$ & 1270 & Fe, As, materials, Fe2O3 & 10.96 \\
$Operation$ & 2439 & heated, sealed, mixed & 21.05 \\
$Property-pressure$ & 401 & air, argon atmosphere, vacuum & 3.46 \\   
$Property-rate$ & 126 & heating rate, cooling rate, 1 K/min & 1.09 \\
$Property-temperature$ & 664 & room temperature, temperature, 900 °C & 5.73 \\
$Property-time$ & 506 & 24 h, 30 min, 2 days & 4.37 \\
$Value$ & 1378 & >99.9\%, stoichiometric amounts, 10 mg  & 11.89 \\
\hline
Overall & 11585 &  & 100.0 \\ 
\bottomrule
\end{tabular}
\caption{Annotated entity mention statistics in the training set.}
\label{tab:NERs}
\end{table*}

\begin{table*}
\centering

\begin{tabular}{c|c|c|c|c}
\textbf{ Relation Label} & \textbf{Count} & \textbf{Frequent mentions} & \textbf{Percentage}\\
\toprule
$Descriptor-of$ & 2796 & (purity, 99.6\%), (Polycrystalline, materials) & 25.02 \\
$Participant-material$ & 2147 & (Pb, melting), (SrCO3, sealed) & 19.21 \\
$Coreference$ & 1171 & (OsO2, powder), (CuO, mixture) & 10.48 \\ 
$Value-of$ & 1737 & (99.99\%, Bi2O3), (50 mg, I2) & 15.54\\
$Condition-of$ & 1547 & (800 °C, heated), (10 hours, held) & 13.84  \\ 
$Next-operation$ & 805 & (kept, heated), (sealed, evacuated) & 7.20 \\ 
$Device-of-operation$ & 637 & (glove box, grinding), (calcined, ground) & 5.70 \\
$Brand-of$ & 336 & (Aldrich, (TPrA)Br), (Alfa Aesar, ZrO2, ) & 3.01 \\
\hline
Overall & 11176 &  & 100.0 \\ 
\bottomrule
\end{tabular}
\caption{Annotated relation pair statistics in the training set.}
\label{tab:REs}
\end{table*}

\end{document}